\documentclass{article}

\PassOptionsToPackage{numbers}{natbib}
\usepackage[preprint]{neurips_2026}
\usepackage[T1]{fontenc}
\usepackage{amsmath,amssymb}
\usepackage{booktabs}
\usepackage{graphicx}
\usepackage{xcolor}
\usepackage{hyperref}
\usepackage{microtype}
\usepackage{url}
\graphicspath{{figures/}{paper/figures/}}

\title{Budgeted Attention Allocation: Cost-Conditioned Compute Control for Efficient Transformers}
\author{%
  Amrit Nidhi\\
  Independent Researcher\\
  \texttt{amrit.nidhi.research@gmail.com}
}
\date{}

\begin{document}
\maketitle

\begin{abstract}
Transformers usually expose one inference cost per trained model, while deployed systems often need multiple cost-quality operating points. We study Budgeted Attention Allocation, a monotone head-gating mechanism conditioned on a requested attention budget. Dense warm-starting is important for stability: on a robust synthetic sequence task, one budgeted model reaches 99.7\% accuracy at 0.303 estimated attention cost and 100.0\% accuracy at 0.504 cost. On held-out AG News with a custom word-level transformer, hard-gate adaptation turns soft cost control into measured single-thread CPU speed, reaching 82.1\% accuracy with 1.28x speedup at budget 0.50. In pretrained BERT-Mini AG News, budgeted structural pruning reaches 87.6\% accuracy with 1.20x speedup at budget 0.50; a validation-ranked zero-shot dense post-hoc structural baseline reaches 86.1\%, and one recovery epoch raises that per-budget specialist to 87.9\%. On DBpedia14, BERT-Mini budgeted gates reach 97.4\% at exact budget 0.50 versus 96.6\% for dense full attention. Static fixed-budget gates and recovered dense specialists remain strong. The contribution is therefore not universal dominance, but a reproducible feasibility study of one controllable checkpoint across budgets that can trade attention cost for accuracy and be converted into measured structural speedups on small CPU benchmarks.
\end{abstract}

\section{Introduction}

Transformer attention \citep{vaswani2017attention} is powerful but expensive. For a sequence of length $n$ and hidden size $d$, dense self-attention has an approximate cost of $O(n^2d)$ per layer. In many practical settings, however, the available compute is not fixed by architecture alone. A deployed model may need to answer cheap requests at low cost, reserve higher compute for difficult inputs, or run under device-specific latency and energy budgets.

This paper explores a cost-conditioned view of attention. Instead of asking only which fixed sparse pattern or compressed model is best, we ask whether a transformer can learn to allocate attention under an explicit budget. The motivating analogy is structural optimization: a bridge does not use material uniformly, but places support where it contributes most to strength per unit cost. Similarly, a transformer should spend attention where it contributes most to predictive performance per unit compute.

The goal is not to outperform every fixed-budget specialist, but to test whether attention cost can be exposed as an inference-time control variable from one checkpoint while retaining competitive accuracy and allowing optional structural materialization.

We make three contributions:

\begin{itemize}
  \item We formulate attention allocation as budget-constrained optimization with an inference-time budget variable.
  \item We implement a lightweight head-gating mechanism and a dense-warm-start training recipe that stabilizes longer-sequence budgeted training.
  \item We evaluate whether one controllable checkpoint across budgets can match the cost-quality behavior of dense attention, post-hoc pruning, validation-ranked dense structural head pruning with and without recovery, and per-budget static gates, including measured single-thread CPU hard-skipping, pretrained BERT-Tiny/BERT-Mini structural pruning, a second DBpedia14 pretrained text task, and the number of fine-tuned models required to support multiple budgets.
\end{itemize}

\section{Related Work}

Efficient transformer research includes structured sparse attention, adaptive span, token pruning, head pruning, and adaptive depth. Sparse attention methods such as Longformer \citep{beltagy2020longformer} and BigBird \citep{zaheer2020bigbird} reduce quadratic cost through predefined patterns. Adaptive attention span \citep{sukhbaatar2019adaptive} learns context length per head. Head pruning work shows that many heads can be removed with limited loss \citep{michel2019sixteen}, while structured pruning methods such as movement pruning \citep{sanh2020movement} and block pruning \citep{lagunas2021block} remove weights, heads, or blocks during fine-tuning. Token pruning methods such as PoWER-BERT \citep{goyal2020powerbert} and DynamicViT \citep{rao2021dynamicvit} remove low-utility tokens during inference. Early-exit methods such as DeeBERT \citep{xin2020deebert}, PABEE \citep{zhou2020bert}, and LayerDrop-style depth control \citep{fan2020layerdrop} adapt depth to input difficulty or deployment constraints.

The idea of one model supporting multiple resource settings also appears in slimmable networks \citep{yu2019slimmable}, Once-for-All networks \citep{cai2020onceforall}, and DynaBERT \citep{hou2020dynabert}. Those methods train width/depth subnetworks; recent adaptive head-budgeting work introduces BudgetFormer, which learns input-dependent head budgets and relevance distributions \citep{faye2026adaptive}. Our setting instead conditions attention-head gates on an externally requested budget so one checkpoint exposes multiple deployment operating points. Direct comparison to width/depth controllers and early-exit policies is left to future work.

Budgeted Attention Allocation is also close in spirit to adaptive computation \citep{graves2016adaptive}: compute should be spent conditionally rather than uniformly. The goal is not to introduce a new fixed sparse pattern. Instead, we expose attention cost as a direct inference-time control variable, so the same trained model can run at multiple price points. This distinguishes the method from static head pruning, where each operating point is typically represented by a separate mask or fine-tuned model; it also distinguishes it from early exit, where the main control variable is depth rather than attention-head allocation.

\section{Method}

Consider a transformer with $L$ layers and $H$ attention heads per layer. Let $B \in (0,1]$ denote the requested attention budget. For the soft controller, we clip the logit input to $\bar{B}=\min(\max(B,\epsilon),1-\epsilon)$ with $\epsilon=10^{-4}$, so the $B=1.00$ endpoint used in evaluation is finite. For each layer $l$ and head $h$, we learn a gate:

\begin{equation}
z_{l,h}(B) = \sigma\left(\frac{a_{l,h} + \operatorname{softplus}(s_{l,h})\operatorname{logit}(\bar{B})}{\tau}\right),
\end{equation}

where $a_{l,h}$ is a learned head logit, $s_{l,h}$ is a learned nonnegative budget sensitivity, $\tau$ is a temperature, and $\sigma$ is the logistic sigmoid. The positive budget sensitivity makes the gate monotonic in $B$: requesting a higher budget increases the probability mass assigned to attention heads. All reported budget controllers use $\tau=1.0$.

For head output $o_{l,h}$, the budgeted output is:

\begin{equation}
\tilde{o}_{l,h} = z_{l,h}(B)o_{l,h}.
\end{equation}

The estimated attention cost fraction is:

\begin{equation}
C(B) = \frac{1}{LH}\sum_{l=1}^{L}\sum_{h=1}^{H} z_{l,h}(B).
\end{equation}

For hard-mask evaluation, we convert soft scores to a hard budget by keeping:

\begin{equation}
k(B)=\max(1,\operatorname{round}(BLH))
\end{equation}

heads across all layers by selecting the $k(B)$ largest soft gate values $z_{l,h}(B)$ and setting the remaining gates to zero. The hard-skipping implementation executes attention only for active heads. Because hard selection is non-differentiable, the custom hard-skip adaptation uses a straight-through top-$k$ estimator and distillation from the soft-gated checkpoint. In pretrained structural adaptation, the controller is frozen and the selected top-$k$ mask is treated as fixed, so no gradient is propagated through top-$k$ selection.

Hard-mask evaluation and structural pruning have different feasibility constraints. A global hard mask can evaluate very low budgets even when some layers receive no active head. Structural pruning is stricter: a materialized BERT model must keep at least one active head in every layer. The reported BERT-Tiny and BERT-Mini structural pruning results use budgets 0.50 and 0.75, which satisfy this per-layer floor. The lower-budget BERT-Mini sweep points, such as requested budget 0.10, are hard-mask evaluations rather than structurally pruned models. The dense post-hoc pruning baseline enforces the same per-layer floor before materializing pruned heads.

For classification with labels $y$, logits $\hat{y}$, and cross-entropy loss $\mathcal{L}_{task}$, we train with:

\begin{equation}
\mathcal{L} = \mathcal{L}_{task}(y,\hat{y}) + \lambda C(B) + \beta \max(0, C(B)-B)^2.
\end{equation}

The first regularizer encourages cheaper attention; the second penalizes budget violations. Unless otherwise stated, training samples budgets uniformly from $[\!0.25,1.00]$, uses $\lambda=0.02$ and $\beta=2.0$, and uses the same distribution during hard-gate adaptation. The robust sequence rescue uses stronger cost pressure ($\lambda=0.05$, $\beta=4.0$), while the DistilBERT pilot uses $\lambda=0.01$ and $\beta=2.0$. Soft gates reduce effective head contribution and define estimated attention cost. Wall-clock savings require structural execution, which we test with the hard-skipping path above.

Hard-gate adaptation updates model weights for one additional epoch under sampled hard budgets. For a sampled budget $B$, let $h(B)$ be the hard top-$k$ mask and $z(B)$ be the soft mask. We use the soft-gated checkpoint with frozen weights $\theta_T$ as a teacher and train student weights $\theta_S$ by minimizing

\begin{equation}
(1-\alpha)\mathcal{L}_{task}(y,f_{\theta_S}(x;h(B))) + \alpha T^2 \operatorname{KL}\left(
\operatorname{softmax}\frac{f_{\theta_T}(x;z(B))}{T}
\;\middle\|\;
\operatorname{softmax}\frac{f_{\theta_S}(x;h(B))}{T}
\right).
\end{equation}

The reported hard-adaptation runs use $\alpha=0.5$ and $T=2.0$. The custom AG News model uses learning rate $3\times10^{-4}$, while the pretrained BERT adaptations use learning rate $2\times10^{-5}$.

\section{Deployment Motivation}

Budget-conditioned attention is useful when inference cost is a first-class constraint. The intended settings are:

\begin{itemize}
  \item multi-tier API serving, where the same model supports low-cost, balanced, and high-quality modes;
  \item mobile and edge inference, where available compute and energy vary by device state;
  \item long-context document classification, where only a small subset of token interactions may be relevant;
  \item batch inference systems, where easy examples can be processed cheaply and hard examples can receive more attention;
  \item task-conditioned routing, where summarization, classification, retrieval, and reasoning may benefit from different attention budgets.
\end{itemize}

\section{Experimental Protocol}

We begin with a synthetic marked-token task. Each sequence contains two markers at random positions; the label depends on whether the values following those markers match. This task is cheap, controlled, and requires sparse evidence extraction across the sequence.

We evaluate:

\begin{itemize}
  \item validation accuracy;
  \item estimated attention cost fraction $C(B)$;
  \item accuracy versus cost Pareto curves across budgets $B \in \{0.25, 0.50, 0.75, 1.00\}$;
  \item learned gate distributions by layer and head.
\end{itemize}

Baselines include dense full attention, post-hoc head pruning from a trained dense model, static learned gates trained for a fixed budget, and budget-conditioned gates. Unless otherwise noted, reported attention cost is estimated by mean gate mass, not measured latency. We use three optimization seeds for the main budgeted/static comparisons. The held-out AG News experiments use seeds 7, 13, and 21; validation data are used only for checkpoint selection, and final accuracy is reported on held-out test examples. The DBpedia14 extension uses the same BERT-Mini protocol with 2,000 training examples, 1,000 validation examples, 1,000 held-out test examples, sequence length 128, and 14 labels.

\section{Results}

\subsection{Robust Sequence Length 64}

Initial sequence length 64 runs were unstable, including for full attention. We therefore calibrated the benchmark until dense full attention was reliable: 8,192 training examples, 2,048 validation examples, and 32 epochs. In a $3\times3$ data-seed by optimization-seed grid, dense full attention averaged 99.4\% validation accuracy, with all nine cells above 95\%.

From-scratch budgeted training did not solve this robust setting. As Table~\ref{tab:seq64-core} shows, the from-scratch budgeted model averaged only 85.1\% accuracy at budget 0.50 because one seed collapsed. Dense warm-starting fixed this failure: initializing from the dense checkpoint and fine-tuning with stronger gate-cost pressure yielded 99.7\% accuracy at 0.303 cost and 100.0\% accuracy at 0.504 cost.

\begin{table}[t]
\centering
\resizebox{\linewidth}{!}{\begin{tabular}{lccccc}
\toprule
Method & Models & Knob & Budget & Cost & Acc. (\%) \\
\midrule
Dense full & 1 & No & 1.00 & $1.000\pm0.000$ & $99.4\pm1.1$ \\
Budgeted from scratch & 1 & Yes & 0.50 & $0.380\pm0.050$ & $85.1\pm25.7$ \\
Warm-start budgeted & 1 & Yes & 0.25 & $0.303\pm0.004$ & $99.7\pm0.3$ \\
Warm-start budgeted & 1 & Yes & 0.50 & $0.504\pm0.002$ & $100.0\pm0.1$ \\
Warm-start static & 3 & No & 0.25 & $0.358\pm0.070$ & $100.0\pm0.0$ \\
Warm-start static & 3 & No & 0.50 & $0.476\pm0.031$ & $100.0\pm0.0$ \\
Post-hoc pruning & 1 + masks & Discrete & 0.50 & $0.500\pm0.000$ & $81.7\pm15.1$ \\
Post-hoc pruning & 1 + masks & Discrete & 0.75 & $0.750\pm0.000$ & $98.8\pm0.8$ \\
\bottomrule
\end{tabular}
}
\caption{Core robust sequence length 64 results. Cost and accuracy report mean $\pm$ standard deviation over three seeds. ``Models'' counts the fine-tuned models needed to support budgets 0.25, 0.50, and 0.75. Static gates are strong when trained per budget; budgeted gates provide one multi-budget model.}
\label{tab:seq64-core}
\end{table}

\begin{figure}[t]
\centering
\includegraphics[width=0.78\linewidth]{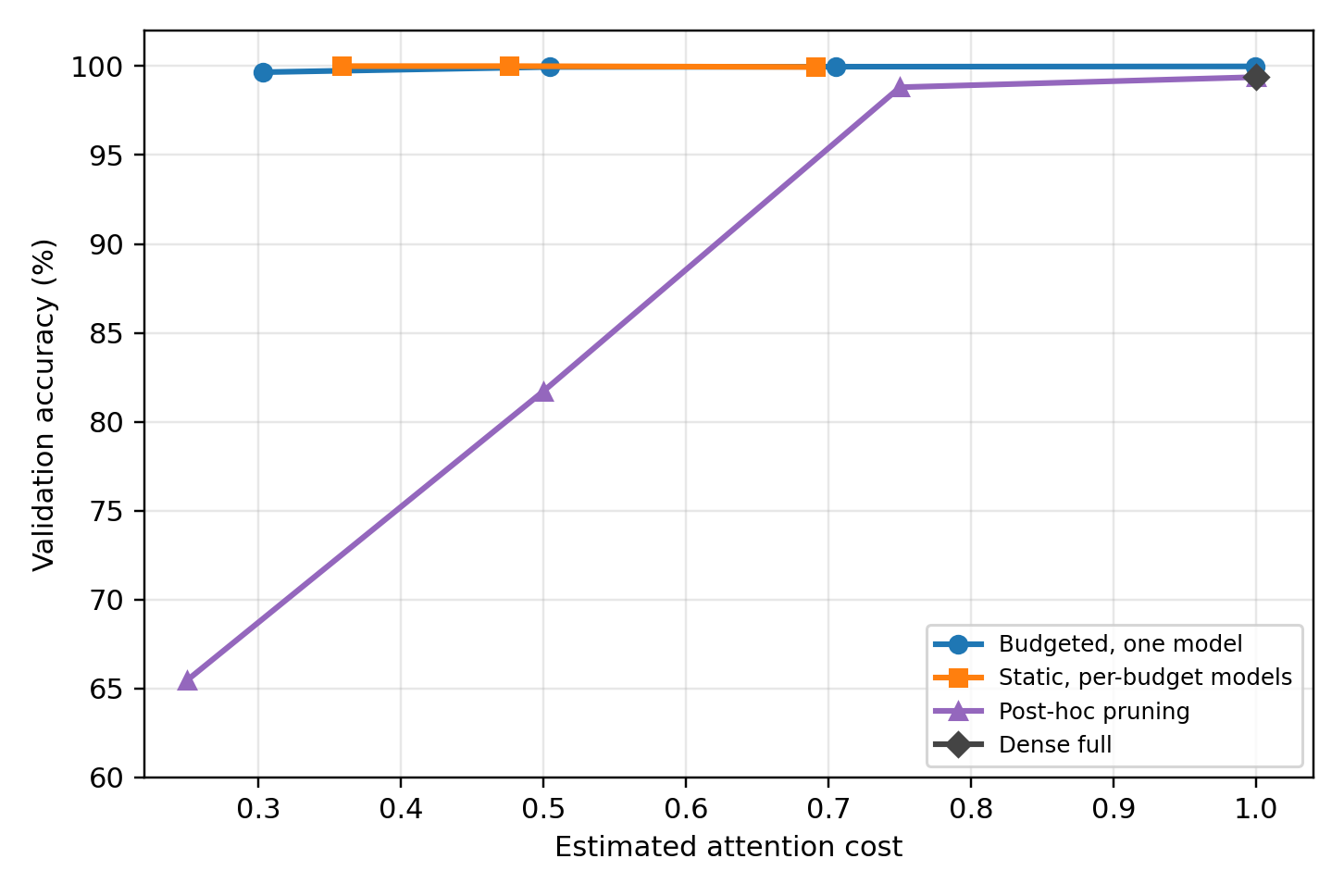}
\caption{Equal-pretraining comparison on robust sequence length 64. Static gates become a strong baseline when initialized from the same dense checkpoint, but require separate fine-tuned models per target budget.}
\label{fig:equal-pretraining}
\end{figure}

\subsection{Controllability}

The equal-pretraining comparison in Figure~\ref{fig:equal-pretraining} makes the story more precise. Warm-start static gates reach 100.0\% accuracy at 0.358 cost for budget 0.25 and 100.0\% at 0.476 cost for budget 0.50. Warm-start budgeted attention reaches 99.7\% at 0.303 cost and 100.0\% at 0.504 cost. Thus, budgeted attention is more cost-efficient at the aggressive low-budget point, while static gates are slightly stronger at the mid-budget point.

The practical distinction is controllability. For $N$ target budgets, warm-start static gates require one dense warm start plus $N$ budget-specific fine-tunes and $N$ deployed specialists. Recovered dense pruning similarly requires separate recovery for each selected budget. Warm-start budgeted gates require one dense warm start, one budgeted fine-tune, and optional hard adaptation, but expose one checkpoint that can be evaluated across many budgets. Table~\ref{tab:model-multiplicity} summarizes this deployment accounting.

\begin{table}[t]
\centering
\resizebox{\linewidth}{!}{\begin{tabular}{lccc}
\toprule
Method & Fine-tune jobs & Artifacts for $N$ budgets & Control \\
\midrule
Dense full & 1 & 1 model & Single cost \\
Dense post-hoc pruning & 1 & $N$ masks/pruned models & Discrete \\
Dense post-hoc + recovery & $1+N$ & $N$ recovered pruned models & Discrete \\
Warm-start static & $1+N$ & $N$ specialists & Discrete \\
Budgeted soft & 2 & 1 model & Continuous estimated cost \\
Budgeted hard-adapt & 3 & 1 model, optional masks & Continuous + structural \\
\bottomrule
\end{tabular}
}
\caption{Deployment accounting for supporting $N$ operating budgets. Fine-tune jobs include the dense warm start where applicable; optional hard masks can be materialized from the same budgeted checkpoint.}
\label{tab:model-multiplicity}
\end{table}

To test whether the budget variable is a genuine control knob, we evaluated the same warm-start budgeted checkpoints at requested budgets from 0.10 to 1.00 in increments of 0.05. Estimated attention cost was monotonic for all three seeds, and accuracy rose smoothly from 91.8\% at 0.157 cost to 99.7\% at 0.303 cost before saturating.

\subsection{Held-Out AG News Text Classification}

We then moved to real text classification on AG News \citep{zhang2015character} using a simple word-level tokenizer, 10,000 training examples, 2,000 validation examples split from the training set, 2,000 held-out test examples from the official test file, and sequence length 128. Checkpoints are selected on the validation split, while Table~\ref{tab:ag-news-small} reports held-out test accuracy. Warm-start budgeted attention reaches 83.47\% accuracy at 0.503 estimated attention cost and 83.87\% at 0.680 cost, compared with 84.00\% for dense attention at full cost. Warm-start static gates remain the strongest per-budget baseline, reaching 84.52\% at 0.259 cost and 84.37\% at 0.529 cost.

\begin{table}[t]
\centering
\resizebox{\linewidth}{!}{\begin{tabular}{lccccc}
\toprule
Method & Models & Knob & Budget & Cost & Acc. (\%) \\
\midrule
Dense full & 1 & No & 1.00 & $1.000\pm0.000$ & $84.0\pm0.4$ \\
Warm-start budgeted & 1 & Yes & 0.25 & $0.325\pm0.006$ & $81.8\pm2.0$ \\
Warm-start budgeted & 1 & Yes & 0.50 & $0.503\pm0.003$ & $83.5\pm0.8$ \\
Warm-start budgeted & 1 & Yes & 0.75 & $0.680\pm0.001$ & $83.9\pm0.8$ \\
Warm-start static & 3 & No & 0.25 & $0.259\pm0.051$ & $84.5\pm0.7$ \\
Warm-start static & 3 & No & 0.50 & $0.529\pm0.036$ & $84.4\pm1.0$ \\
Warm-start static & 3 & No & 0.75 & $0.736\pm0.021$ & $84.2\pm1.3$ \\
Post-hoc pruning & 1 + masks & Discrete & 0.50 & $0.500\pm0.000$ & $78.0\pm0.2$ \\
Post-hoc pruning & 1 + masks & Discrete & 0.75 & $0.750\pm0.000$ & $81.8\pm1.3$ \\
\bottomrule
\end{tabular}
}
\caption{Held-out AG News small result with validation-only checkpoint selection. Cost and accuracy report mean $\pm$ standard deviation over three seeds. Static gates remain strongest per fixed budget, while budgeted gates provide a single controllable model across budgets.}
\label{tab:ag-news-small}
\end{table}

\subsection{Hard-Gate Adaptation and Latency}

The results above report mean gate mass as estimated attention cost. To test whether this proxy can become real speed, we added an inference-time hard head-skipping path. For binary gates, the implementation computes query, key, and value projections only for active heads, applies attention only for those heads, and uses only the corresponding output-projection columns. We then fine-tuned each budgeted checkpoint with straight-through top-$k$ gates and, in the main setting, distillation from the original soft-gated checkpoint. This is still a naive PyTorch implementation rather than a fused kernel.

Without adaptation, hard skipping is too damaging at aggressive budgets: budget 0.50 reaches only 74.8\% held-out accuracy. One epoch of hard-gate exposure with distillation from the original soft-gated checkpoint improves this to 82.1\%, so we use that cheaper one-epoch recipe for the latency evaluation.

Table~\ref{tab:ag-news-latency} reports single-thread CPU latency on 2,000 held-out AG News examples, averaged over three seeds and five repeats. Soft gates preserve accuracy but do not improve latency because every head is still computed. Hard-gate adaptation substantially improves the hard-skip regime: budgeted hard skipping reaches 82.1\% held-out accuracy with 1.28x speedup at budget 0.50, and 83.9\% accuracy with 1.09x speedup at budget 0.75. Static hard skipping remains competitive, reaching 82.8\% accuracy with 1.25x speedup at budget 0.50.

\begin{table}[t]
\centering
\resizebox{\linewidth}{!}{\begin{tabular}{llccccc}
\toprule
Method & Eval & Budget & Cost & Acc. (\%) & Latency ms & Speedup \\
\midrule
Dense & dense & 1.00 & 1.000 & $84.0\pm0.4$ & $2323.5\pm47.2$ & 1.00 \\
Budgeted & soft & 0.50 & 0.509 & $84.6\pm1.0$ & $2517.7\pm159.6$ & 0.93 \\
Budgeted & hard skip & 0.50 & 0.500 & $82.1\pm1.7$ & $1836.5\pm238.3$ & 1.28 \\
Budgeted & hard skip & 0.75 & 0.750 & $83.9\pm0.3$ & $2134.8\pm197.2$ & 1.09 \\
Static & hard skip & 0.50 & 0.500 & $82.8\pm0.7$ & $1873.8\pm183.5$ & 1.25 \\
Static & hard skip & 0.75 & 0.750 & $82.8\pm1.0$ & $2047.1\pm37.4$ & 1.13 \\
\bottomrule
\end{tabular}
}
\caption{AG News single-thread CPU latency on 2,000 held-out examples. Accuracy and latency report mean $\pm$ standard deviation over three seeds. Speedup averages per-seed dense/method latency ratios rather than dividing the displayed aggregate latencies. Soft gates estimate cost but do not skip computation; the soft row evaluates the hard-adapted checkpoint and therefore differs from the validation-selected checkpoint in Table~\ref{tab:ag-news-small}. Hard skip uses binary gates to skip inactive heads after one epoch of hard-gate adaptation for the budgeted model.}
\label{tab:ag-news-latency}
\end{table}

\subsection{Pretrained BERT-Tiny and BERT-Mini Head Pruning}

To test whether the result survives beyond a custom word-level transformer, we first ran a pretrained BERT-Tiny extension using the Hugging Face Transformers library \citep{devlin2019bert,wolf2020transformers}. Across three seeds, hard-adapted budgeted BERT-Tiny reaches 84.5\% accuracy at exact hard budget 0.50, compared with 85.7\% for a four-epoch dense model. Structural \texttt{prune\_heads} materialization preserves that 84.5\% accuracy while improving median single-thread CPU latency from 1348.9 ms to 924.6 ms, a 1.45x speedup.

To check that the result was not limited to the two-layer BERT-Tiny model, we repeated the same pretrained protocol with BERT-Mini (\texttt{google/bert\_uncased\_L-4\_H-256\_A-4}), a 4-layer encoder with 4 heads per layer and 11.2M parameters. Table~\ref{tab:bert-mini-path-b} shows that hard-adapted budgeted BERT-Mini remains competitive with dense and static baselines: at exact hard budget 0.50 it reaches 87.6\% accuracy, compared with 87.2\% for dense full attention and 86.7\% for the static hard specialist.

We then evaluated the saved BERT-Mini budgeted checkpoints at requested budgets from 0.10 to 1.00 in increments of 0.05. For all three seeds, both the soft mean-gate cost curve and hard top-$k$ cost curve were monotonic. Representative soft points are 84.2\% accuracy at 0.194 cost for requested budget 0.10, 86.9\% at 0.498 cost for budget 0.50, and 87.1\% at 0.998 cost for budget 1.00. Representative hard points are 87.2\% at exact 0.250 cost, 87.6\% at exact 0.500 cost, and 87.4\% at exact 0.750 cost. This directly supports the pretrained inference-time budget knob, while showing that accuracy saturates quickly on this small AG News split.

\begin{table}[t]
\centering
\resizebox{0.74\linewidth}{!}{\begin{tabular}{lccc}
\toprule
Curve & Budget & Cost & Acc. \\
\midrule
Soft & 0.10 & $0.194 \pm 0.002$ & $84.2 \pm 3.6$ \\
Soft & 0.25 & $0.327 \pm 0.005$ & $86.3 \pm 1.2$ \\
Soft & 0.50 & $0.498 \pm 0.007$ & $86.9 \pm 1.4$ \\
Soft & 0.75 & $0.672 \pm 0.008$ & $87.0 \pm 1.2$ \\
Soft & 1.00 & $0.998 \pm 0.000$ & $87.1 \pm 1.5$ \\
Hard & 0.10 & 0.125 & $81.5 \pm 6.7$ \\
Hard & 0.25 & 0.250 & $87.2 \pm 1.6$ \\
Hard & 0.50 & 0.500 & $87.6 \pm 1.2$ \\
Hard & 0.75 & 0.750 & $87.4 \pm 0.7$ \\
Hard & 1.00 & 1.000 & $87.1 \pm 1.5$ \\
\bottomrule
\end{tabular}
}
\caption{Pretrained BERT-Mini continuous budget sweep over three seeds. The full sweep evaluates 19 requested budgets from 0.10 to 1.00; both soft and hard cost curves are monotonic for every seed.}
\label{tab:bert-mini-budget-sweep}
\end{table}

Table~\ref{tab:bert-mini-pruning} shows that structural pruning at budget 0.50 improves median single-thread CPU latency from 4644.5 ms to 3892.9 ms, a 1.20x speedup. The corresponding compute analysis removes 4.7\% of total parameters, 50\% of head-specific attention parameters and estimated attention MACs, and 19.2\% of estimated transformer-layer MACs at sequence length 128.

We also compare against direct post-hoc dense pruning baselines. For each seed, we score each dense BERT-Mini head by the validation-loss increase caused by individually masking that head, keep the highest-scoring heads under the same hard budget with a per-layer floor, and then materialize the resulting model with \texttt{prune\_heads}. Table~\ref{tab:bert-mini-dense-pruning} shows that at budget 0.50, budgeted structural pruning has higher mean accuracy than zero-shot dense post-hoc structural pruning (87.6\% versus 86.1\%) at comparable measured speed (1.20x for both), but the seed-level standard deviations overlap and this should be read as a favorable mean in a small run rather than a significance claim. We therefore run an adaptation-matched recovery baseline: after dense structural pruning, each budget-specific specialist receives one recovery epoch on the same 2,000-example train split using AdamW with learning rate $2\cdot10^{-5}$ and dense-teacher distillation ($\alpha=0.5$, temperature 2.0). Recovery raises dense post-hoc accuracy to 87.9\% at budget 0.50 and 88.1\% at budget 0.75, with measured speedups of 1.24x and 1.12x after saving and reloading the recovered pruned model before timing. Thus recovered dense pruning is stronger at fixed budgets; the budgeted model's advantage is the single controllable checkpoint rather than fixed-budget dominance.

\begin{table}[t]
\centering
\begin{tabular}{llccc}
\toprule
Method & Eval & Budget & Cost & Acc. \\
\midrule
Dense & full & 1.00 & 1.000 & $87.2 \pm 1.1$\% \\
Budgeted + hard adapt & hard top-$k$ & 0.50 & 0.500 & $87.6 \pm 1.2$\% \\
Budgeted + hard adapt & hard top-$k$ & 0.75 & 0.750 & $87.4 \pm 0.7$\% \\
Static specialist & hard top-$k$ & 0.50 & 0.500 & $86.7 \pm 1.0$\% \\
Static specialist & hard top-$k$ & 0.75 & 0.750 & $87.8 \pm 0.9$\% \\
\bottomrule
\end{tabular}

\caption{Pretrained BERT-Mini AG News results over three seeds. Values report mean $\pm$ standard deviation over seeds. The budgeted model uses one controllable checkpoint; static rows use separate fixed-budget specialists.}
\label{tab:bert-mini-path-b}
\end{table}

\begin{table}[t]
\centering
\resizebox{\linewidth}{!}{\begin{tabular}{llcccc}
\toprule
Variant & Budget & Cost & Acc. & Median ms & Speedup \\
\midrule
Unpruned hard-adapt model & 1.00 & 1.000 & $87.1 \pm 1.5$\% & $4644.5 \pm 183.6$ & 1.00 \\
Hard head mask & 0.50 & 0.500 & $87.6 \pm 1.2$\% & $4724.0 \pm 72.1$ & $0.98 \pm 0.03$ \\
Structural pruning & 0.50 & 0.500 & $87.6 \pm 1.2$\% & $3892.9 \pm 183.2$ & $1.20 \pm 0.10$ \\
Hard head mask & 0.75 & 0.750 & $87.4 \pm 0.7$\% & $4807.9 \pm 59.8$ & $0.97 \pm 0.05$ \\
Structural pruning & 0.75 & 0.750 & $87.4 \pm 0.7$\% & $4299.6 \pm 95.3$ & $1.08 \pm 0.04$ \\
\bottomrule
\end{tabular}
}
\caption{Pretrained BERT-Mini single-thread CPU latency with hard-adapted budgeted gates. Values report mean $\pm$ standard deviation over seed-level medians, except accuracy which is mean $\pm$ standard deviation over seeds. Structural pruning gives measured speedup where hard masking alone does not.}
\label{tab:bert-mini-pruning}
\end{table}

\begin{table}[t]
\centering
\resizebox{\linewidth}{!}{\begin{tabular}{lccccc}
\toprule
Method & Budget & Cost & Acc. & Median ms & Speedup \\
\midrule
Dense BERT-Mini & 1.00 & 1.00 & $87.2 \pm 1.1$ & $4678.1 \pm 70.5$ & $1.00 \pm 0.00$ \\
Budgeted structural & 0.50 & 0.50 & $87.6 \pm 1.2$ & $3892.9 \pm 183.2$ & $1.20 \pm 0.10$ \\
Dense post-hoc structural & 0.50 & 0.50 & $86.1 \pm 2.7$ & $3688.8 \pm 187.2$ & $1.20 \pm 0.03$ \\
Dense post-hoc + recovery & 0.50 & 0.50 & $87.9 \pm 1.2$ & $3570.0 \pm 92.0$ & $1.24 \pm 0.07$ \\
Budgeted structural & 0.75 & 0.75 & $87.4 \pm 0.7$ & $4299.6 \pm 95.3$ & $1.08 \pm 0.04$ \\
Dense post-hoc structural & 0.75 & 0.75 & $87.5 \pm 0.8$ & $3986.6 \pm 153.6$ & $1.11 \pm 0.06$ \\
Dense post-hoc + recovery & 0.75 & 0.75 & $88.1 \pm 1.2$ & $3944.9 \pm 94.4$ & $1.12 \pm 0.06$ \\
\bottomrule
\end{tabular}
}
\caption{Direct BERT-Mini comparison between budgeted structural pruning and validation-ranked dense post-hoc structural pruning. Both methods use the same held-out test slices, single-thread CPU timing protocol, and Hugging Face \texttt{prune\_heads} materialization. Speedups average per-seed dense/method ratios from the corresponding timing run rather than dividing the displayed aggregate medians. Recovered dense rows use one budget-specific recovery epoch and are timed after saving and reloading the pruned model.}
\label{tab:bert-mini-dense-pruning}
\end{table}

A Colab Pro+ A100 sanity check with the same three BERT-Mini AG News seeds finds that these small structural head removals do not materially accelerate Hugging Face eager GPU inference: budgeted structural pruning gives only about 1.02x median speedup at budget 0.50 and 0.99x at budget 0.75, with dense post-hoc pruning similarly near 1.0x. We therefore keep the main latency claim CPU-specific and treat structural speedups as hardware- and kernel-dependent.

To address dataset breadth, we also ran the same BERT-Mini recipe on DBpedia14 \citep{zhang2015character}. Table~\ref{tab:dbpedia-bert-mini} shows that dense full attention reaches 96.6\% accuracy, while budgeted hard-adapted gates reach 97.4\% at exact budget 0.50 and 97.5\% at exact budget 0.75. A 19-point sweep is monotonic for all three seeds; representative soft points are 94.6\% at 0.226 cost for requested budget 0.10, 97.0\% at 0.503 cost for budget 0.50, and 97.6\% at 0.995 cost for budget 1.00. This second-task result strengthens the controllability evidence beyond AG News; DBpedia structural latency and dense-pruning comparisons remain seed-7 pilots and are not used for broad speed claims.

\begin{table}[t]
\centering
\resizebox{0.82\linewidth}{!}{\begin{tabular}{llccc}
\toprule
Method & Eval & Budget & Cost & Acc. \\
\midrule
Dense & full & 1.00 & 1.000 & $96.6 \pm 0.6$\% \\
Budgeted + hard adapt & hard top-$k$ & 0.50 & 0.500 & $97.4 \pm 0.8$\% \\
Budgeted + hard adapt & hard top-$k$ & 0.75 & 0.750 & $97.5 \pm 0.4$\% \\
Static specialist & hard top-$k$ & 0.50 & 0.500 & $95.2 \pm 1.2$\% \\
Static specialist & hard top-$k$ & 0.75 & 0.750 & $94.8 \pm 2.6$\% \\
\bottomrule
\end{tabular}
}
\caption{Pretrained BERT-Mini DBpedia14 results over three seeds. Values report mean $\pm$ standard deviation over seeds on a sampled held-out test slice.}
\label{tab:dbpedia-bert-mini}
\end{table}

A post-hoc gate-ranking analysis finds a stable high-priority core rather than unrelated per-budget rankings: between budgets 0.25 and 0.75, Spearman correlation is 0.91$\pm$0.01 on AG News and 0.84$\pm$0.04 on DBpedia14, with full retention of low-budget top-$k$ heads at the higher budget.

As a larger-encoder cost-control pilot, we also ran a DistilBERT-base experiment \citep{sanh2019distilbert} with 1,000 AG News training examples, 250 validation examples, 500 held-out test examples, and seeds 7, 13, and 21. This pilot uses estimated head-mask cost rather than structural pruning, so it is not a latency claim. Table~\ref{tab:distilbert-pilot} includes an equal-training dense-continued baseline: dense DistilBERT reaches 86.5\% after the initial epoch and 87.4\% after the same additional epoch budgeted/static methods receive. Warm-start budgeted gates remain competitive, reaching 88.1\% at exact 0.500 estimated cost and 88.4\% at 0.682 estimated cost; dense post-hoc pruning is also strong, reaching 88.0\% at budget 0.75. To reduce single-seed risk, we additionally ran a larger-subset Colab A100 check with 5,000 training, 1,000 validation, 1,000 held-out test examples, and the same three seeds. Dense-continued reaches 91.0$\pm$0.4\% at full cost; warm-start budgeted reaches 90.7$\pm$1.2\% at exact 0.500 cost and 90.5$\pm$1.3\% at 0.682 cost; static specialists reach 90.6$\pm$0.4\% and 90.4$\pm$0.7\%; and post-hoc pruning reaches 90.2$\pm$0.9\% and 90.3$\pm$1.1\% at budgets 0.50 and 0.75. Thus the DistilBERT evidence supports competitive estimated cost control on a larger encoder, but it is not evidence for universal dominance or structural speed.

\begin{table}[t]
\centering
\begin{tabular}{lccc}
\toprule
Method & Budget & Cost & Acc. \\
\midrule
Dense DistilBERT & 1.00 & 1.000 & $86.5 \pm 2.2$ \\
Dense DistilBERT continued & 1.00 & 1.000 & $87.4 \pm 1.4$ \\
Warm-start budgeted & 0.50 & $0.500 \pm 0.000$ & $88.1 \pm 1.3$ \\
Warm-start budgeted & 0.75 & $0.682 \pm 0.000$ & $88.4 \pm 0.9$ \\
Warm-start static & 0.50 & $0.504 \pm 0.016$ & $87.6 \pm 1.4$ \\
Dense post-hoc prune & 0.50 & 0.500 & $87.1 \pm 0.6$ \\
Dense post-hoc prune & 0.75 & 0.750 & $88.0 \pm 1.4$ \\
\bottomrule
\end{tabular}

\caption{Larger-encoder DistilBERT-base AG News pilot over three seeds. This table reports estimated head-mask cost only; it does not claim structural latency speedup.}
\label{tab:distilbert-pilot}
\end{table}

\section{Reproducibility}

All experiments except the final A100 sanity checks are CPU-scale experiments implemented in PyTorch. The synthetic suite uses 8,192 training and 2,048 validation examples; the custom AG News study uses 10,000 training, 2,000 validation, and 2,000 held-out test examples. Pretrained BERT-Tiny/BERT-Mini AG News and DBpedia14 use 2,000 training, 1,000 validation, and 1,000 held-out test examples over seeds 7, 13, and 21; DBpedia CSVs are prepared locally from 3,000 training-file rows and 1,000 test-file rows, with raw CSVs and Parquet caches excluded from the reproduction package. Continuous BERT-Mini sweeps evaluate saved checkpoints at 19 requested budgets from 0.10 to 1.00. Pretrained timing uses pre-tokenized materialized batches, \texttt{torch.no\_grad()}, \texttt{eval()} mode, Python 3.11.5, PyTorch 2.8.0, Transformers 4.54.0, macOS 15.6.1 on arm64, and one PyTorch intra/inter-op thread on an eight-core CPU machine with 8GB memory. Training and latency batch sizes are 64 for the custom AG News model, 16 for BERT-Tiny, 8 for BERT-Mini/DBpedia14, and 16 for DistilBERT. Recorded wall-clock logs are about 37 minutes for the synthetic suite, 44 minutes for custom AG News, 39 minutes for AG News BERT-Mini, roughly 40 minutes for DBpedia BERT-Mini, and 47 minutes for the three-seed DistilBERT pilot; evaluation-only sweep and pruning jobs take about 5--20 minutes, and the dense-pruning recovery check takes about 15--25 minutes. The larger-subset DistilBERT check used a Colab Pro+ A100 GPU, the same 5,000/1,000/1,000 split over three seeds, and completed the logged training/evaluation stages in under 10 minutes after setup and downloads; the A100 BERT-Mini latency check used Python 3.12.13, PyTorch 2.10.0+cu128, Transformers 4.54.0, 10 warmup passes, and 30 timed repeats. The reproduction package includes exact commands, configs, tests, license notes, and the ordered driver \texttt{reproduce\_neurips\_key\_results.py}.

\section{Discussion}

The current evidence does not support the broad claim that budgeted attention universally dominates static pruning. It supports a narrower and more useful claim: dense-warm-start budgeted attention provides competitive cost-quality behavior from one controllable checkpoint across budgets, while the strongest static and recovered dense-pruning baselines require separate per-budget fine-tuning or materialization. The held-out AG News and DBpedia results strengthen the cost-quality direction on real text, and the hard-gate adaptation results show that estimated savings can translate into measured speed when implemented as hard skipping or structural pruning. The BERT-Mini dense-pruning baseline further clarifies the contribution: budgeted structural pruning has a higher mean accuracy than validation-ranked zero-shot dense post-hoc pruning at the aggressive 0.50 budget in this small three-seed AG News run, but a one-epoch recovered dense specialist has the highest fixed-budget mean accuracy at both 0.50 and 0.75. Recovered dense pruning is therefore the strongest choice when the deployment budget is known in advance and only one operating point must be served. Budgeted attention addresses a different deployment regime: a single checkpoint must support multiple budgets at inference time without training, selecting, validating, and serving a separate recovered artifact for each target cost.

Training-cost accounting should be separated from deployment accounting. A dense baseline requires one dense fine-tune and one deployed model. Static gates require the dense warm start plus one extra fine-tune per target budget, producing one specialist per operating point. Recovered dense pruning similarly adds one recovery fine-tune per target budget after head scoring and structural materialization. Budgeted attention requires the dense warm start, one budgeted fine-tune, and optional one-epoch hard adaptation, producing one controllable checkpoint that can be evaluated across many budgets and optionally materialized into hard masks. The method is therefore not claimed to reduce total training compute; its present advantage is reducing the number of budget-specific deployment artifacts and exposing a continuous cost knob.

Efficiency improvements may reduce inference cost and energy use for benign deployments, but cheaper inference can also lower the cost of harmful applications such as spam, surveillance, or disinformation systems. Because this work studies a general efficiency mechanism rather than an application-specific model, deployment safeguards depend on the downstream use case.

\section{Limitations}

The initial implementation uses soft gates during most training, so estimated attention-cost reductions do not automatically imply wall-clock speedups. The hard-gate adaptation result is encouraging but limited: the meaningful structural speedups are single-thread CPU results on small encoders, and the A100 sanity check shows that GPU acceleration is not automatic under eager Hugging Face execution without fused kernels or hardware-specific sparse attention support. The DistilBERT-base pilot broadens model scale for estimated cost control but does not yet materialize structural latency speedups. The pretrained text experiments use sampled AG News and DBpedia14 subsets rather than full benchmark training, and the DBpedia structural-latency/dense-pruning comparison is currently only a seed-7 pilot. The adaptation ablation is also small, so it should be read as evidence for the recipe rather than as a final training study. The synthetic task is diagnostic rather than sufficient evidence for broad usefulness. The robust sequence length 64 results also show that from-scratch budgeted training is seed-sensitive; dense warm-starting fixes the tested failure but increases total training compute. Finally, equal-pretraining static gates and post-hoc pruning are strong, so comparisons should report not only accuracy and estimated cost but also the number of budget-specific models required.

\bibliographystyle{plainnat}
\bibliography{references}

\end{document}